\documentclass[10pt,twocolumn,letterpaper]{article}

\usepackage{3dv}
\usepackage{times}
\usepackage{epsfig}
\usepackage{graphicx}
\usepackage{amsmath}
\usepackage{amssymb}

\usepackage{color}
\usepackage{soul}
\usepackage{url}
\usepackage[utf8]{inputenc}
\usepackage{graphicx}
\usepackage{amsthm}
\usepackage{booktabs}
\usepackage{algorithm}
\usepackage{algorithmic}
\usepackage{mathtools}
\usepackage{subcaption}
\usepackage{wrapfig}
\usepackage{booktabs}
\usepackage{blindtext}
\usepackage[T1]{fontenc}
\usepackage[font=small,labelfont=bf,tableposition=top]{caption}
\usepackage[pagebackref=false,breaklinks=true,letterpaper=true,colorlinks,bookmarks=false]{hyperref}

\threedvfinalcopy 

\def\threedvPaperID{182} 

\ifthreedvfinal\fi
\begin{document}

\title{PeeledHuman: Robust Shape Representation for Textured 3D Human Body Reconstruction}

\author{Sai Sagar Jinka\\
\and
Rohan Chacko\\
\and
Avinash Sharma\\
\and
P.J. Narayanan\\
\and
Centre for Visual Information Technology, IIIT Hyderabad, India\\
{\tt\small \{jinka.sagar@research.,rohan.chacko@students., asharma@, pjn@\}iiit.ac.in}
}

\maketitle
\thispagestyle{empty}

\begin{abstract}
    We introduce PeeledHuman - a novel shape representation of the human body that is robust to self-occlusions. PeeledHuman encodes the human body as a set of Peeled Depth and RGB maps in 2D, obtained by performing ray-tracing on the 3D body model and extending each ray beyond its first intersection. This formulation allows us to handle self-occlusions efficiently compared to other representations. Given a monocular RGB image, we learn these Peeled maps in an end-to-end generative adversarial fashion using our novel framework - PeelGAN. We train PeelGAN using a 3D Chamfer loss and other 2D losses to generate multiple depth values per-pixel and a corresponding RGB field per-vertex in a dual-branch setup. In our simple non-parametric solution, the generated Peeled Depth maps are back-projected to 3D space to obtain a complete textured 3D shape. The corresponding RGB maps provide vertex-level texture details. We compare our method with current parametric and non-parametric methods in 3D reconstruction and find that we achieve state-of-the-art-results. We demonstrate the effectiveness of our representation on publicly available BUFF and MonoPerfCap datasets as well as loose clothing data collected by our calibrated multi-Kinect setup.
\end{abstract}

\section{Introduction}

    Reconstruction of a textured 3D model of the human body from images is a pivotal problem in computer vision and graphics.
    It has widespread applications in the entertainment industry, e-commerce, health-care, and AR/VR platforms. Traditional methods for 3D body reconstruction used voxel carving, triangulation, or structured lighting approaches~\cite{bogo2017dynamic,VlasicMIT2008} that require multi-view images captured from calibrated setups.
    Recent advancements in deep learning have renewed interest in this domain with the focus on a more challenging variant of the problem: monocular 3D reconstruction, which inherently is an ill-posed problem.
    This is particularly challenging as the geometry of non-rigid human shapes varies over time, yielding a large space of complex articulated body poses and shape variations. Monocular reconstruction imposes several other challenges such as self-occlusions, obstructions due to free-form clothing, and significant viewpoint variations.
   
    \begin{figure} 
    \begin{center}
        \includegraphics[width=\linewidth]{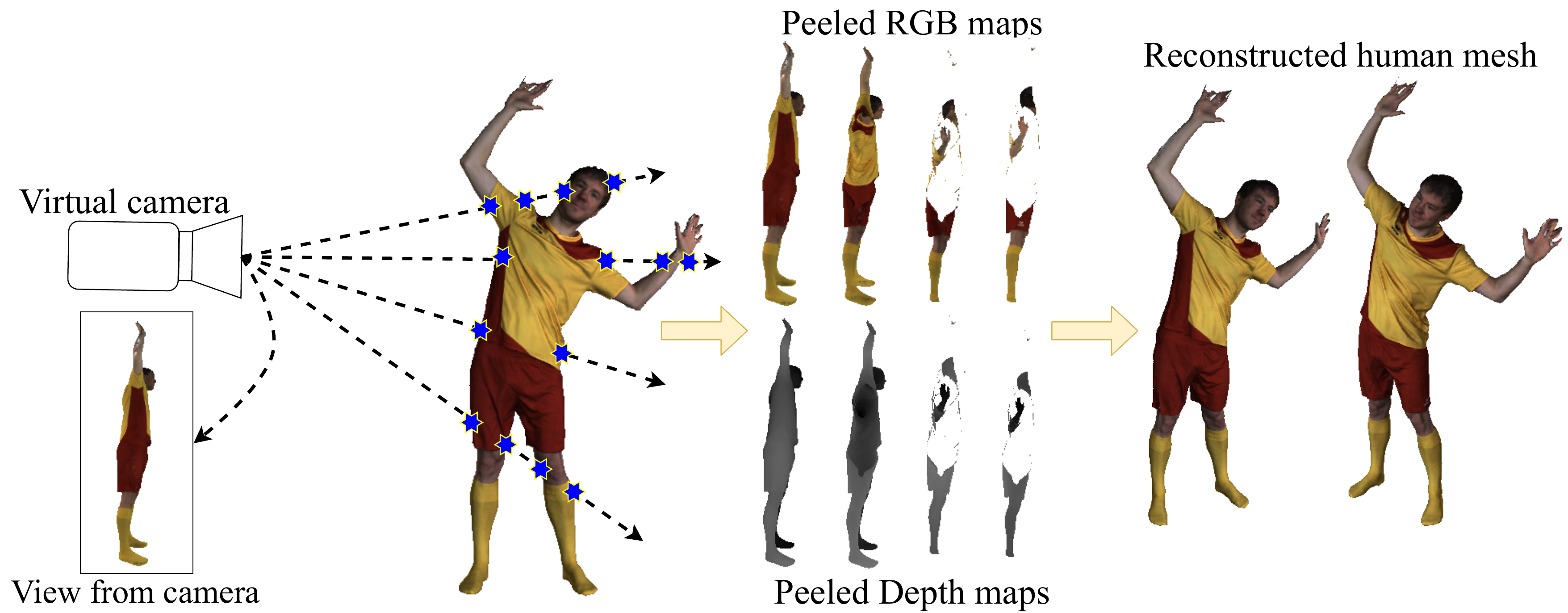}
    \end{center}
    \caption{\textbf{PeeledHuman}. Our proposed representation  encodes a human body as a set of {\em Peeled Depth} \& {\em RGB maps} from a given view. These maps are back-projected to 3D space in the camera coordinate frame to recover the 3D human body.}
    \label{fig:Motivation}
    \end{figure}

    Existing deep-learning solutions for {\em monocular} 3D human reconstruction can be broadly categorized into two classes. The first class of model-based approaches (e.g., ~\cite{hmr,omran2018neural}) attempt to fit a parametric body representation, like the SMPL~\cite{SMPL:2015,pavlakos2019expressive}, to recover the 3D surface model. 
    Such model-based methods efficiently approximate the shape and pose of the underlying naked body but fail to reconstruct fine surface texture details of the body and the wrapped clothing. Parametric SMPL models have been extended to include clothing details like in~\cite{patel2020tailornet, alldieck2019tex2shape}.
    Another approach by~\cite{guler2018densepose} predicts a UV map for every foreground pixel to generate texture over an SMPL model. However, it does not account for large clothing deformations. 
    
    The second class of model-free approaches does not assume any parametric model of the body. One set of model-free approaches employ volumetric regression, a natural extension of 2D convolutions, for human body recovery from a monocular image~\cite{varol18_bodynet,VenkatSS18}. However, volumetric regression is known to be memory intensive and computationally inefficient as it involves redundant 3D convolutions on empty voxels. Additionally, this memory-intensive behavior restricts the ability to learn detailed surface geometry.  

    The recent works in this direction include MouldingNet~\cite{gabeur2019moulding}, PIFu~\cite{saito2019pifu}, and its follow-up work PIFuHD~\cite{saito2020pifuhd}. PIFu proposes a deep network that learns an implicit function to recover 3D human models under loose clothing. More precisely, they compute local per-pixel feature vectors on an inference image and a specified z-depth along the outgoing camera ray from each pixel to learn an implicit function that can classify whether a 3D point corresponding to this z-depth is inside or outside the body surface. However, this requires sampling multiple 3D points from the canonical 3D volume and testing for each of them independently. Such sampling adds redundancy at inference time as a large number of points inside as well as outside the 3D body surface are tested. Instead, identifying the 3D points on the surface is more efficient for recovering the surface geometry.  
    On the other hand, MouldingNet~\cite{gabeur2019moulding} proposes to recover 3D body models by performing a pixel-wise regression of two {\em independent} depth maps (visible and hidden). This is similar to generating depth maps captured by two RGBD virtual cameras separated by $180^\circ$ along z-axis. Although such pixel-wise regression is computationally more efficient as compared to PIFu and can model arbitrary surface topology, it still fails to handle self-occlusions. To summarize, model-based methods cannot reconstruct highly textured clothed subjects with arbitrary shape topologies. On the other hand, existing model-free approaches are either computationally intensive or unable to handle large self-occlusions.
   
    In this paper, we tackle the problem of textured 3D human reconstruction from a single RGB image by introducing a novel shape representation, shown in Figure \ref{fig:Motivation}. 
    Our proposed solution derives inspiration from the classical ray tracing approach in computer graphics. We estimate a fixed number of ray intersection points with the human body surface in the canonical view volume for every pixel in an image, 
    yielding a multi-layered shape representation called {\em PeeledHuman}. PeeledHuman encodes a 3D shape as a set of depth maps called hereinafter as {\em Peeled Depth maps}. We further extend this layered representation to recover texture by capturing a discrete sampling of the continuous surface texture called hereinafter as {\it Peeled RGB maps}. Such a layered representation of the body shape addresses severe self-occlusions caused by complex body poses and viewpoint variations. Our representation is similar to \textit{depth peeling} used in computer graphics for order-independent transparency. The proposed shape representation allows us to recover multiple 3D points that project to the same pixel in the 2D image plane (see Figure \ref{fig:Motivation}), thereby overcoming the limitation of handling self-occlusions in MouldingNet. This solution is also more efficient than PIFu at both training and inference time as it simultaneously (globally) predicts and regresses to a fixed set of Peeled Depth \& RGB maps for an input monocular image.  It is important to note that our representation is not restricted only to human body models but can generalize well to any 3D shapes/scenes, given specific training data prior.  

\begin{figure*}[t!] 
    \begin{center}
        \includegraphics[width=0.95\linewidth]{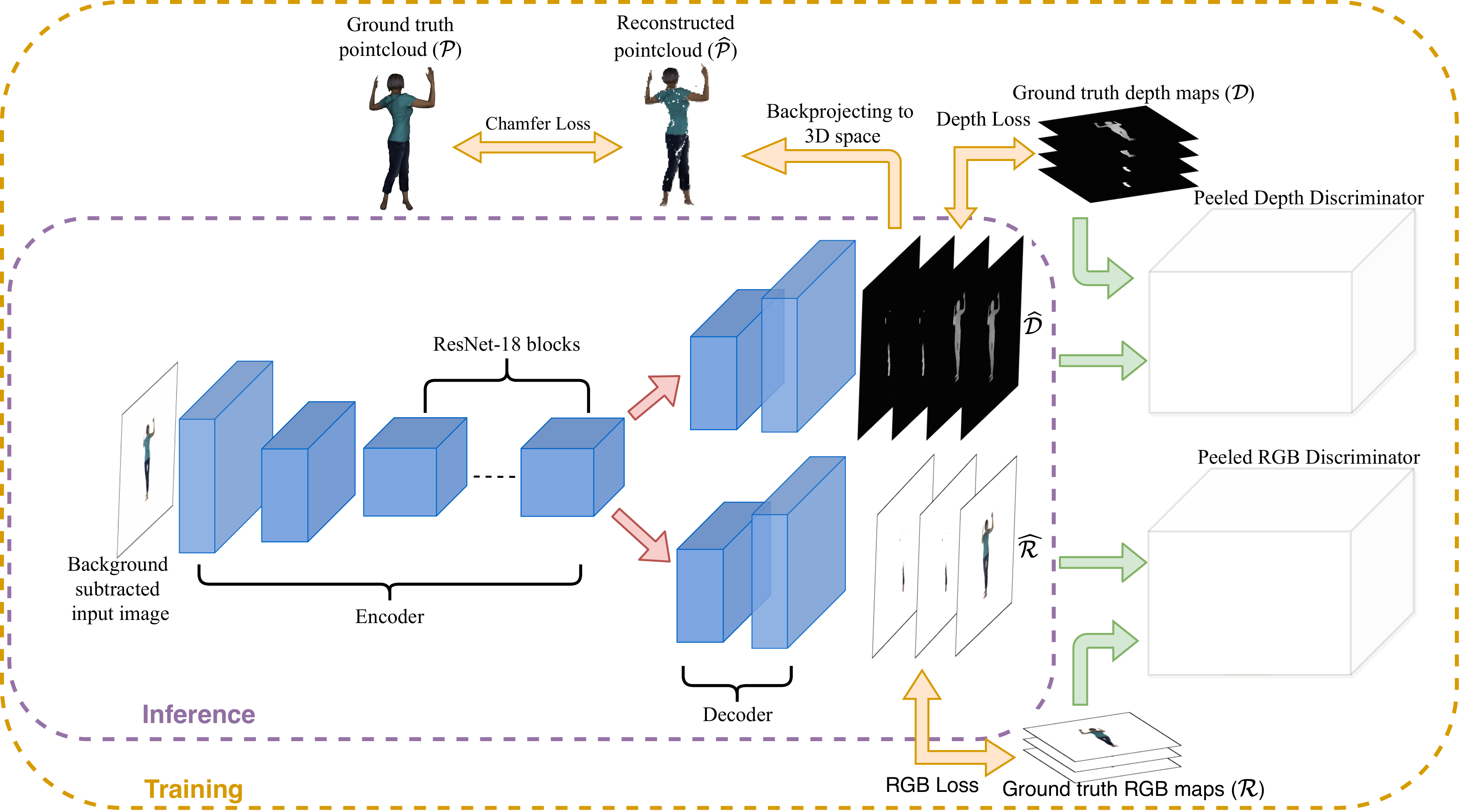}\\[1em]   
    \end{center}
    \caption{\textbf{PeelGAN overview}: The dual-branch network generates Peeled Depth ($\widehat{\mathcal{D}}$) and RGB ($\widehat{\mathcal{R}}$) maps from an input image. The generated maps are each fed to a discriminator: one for RGB and one for Depth maps. The generated maps are back-projected to obtain the 3D human body represented as a point cloud  ($\widehat{\mathcal{P}}$) in the camera coordinate frame. We employ a Chamfer loss between the reconstructed point cloud and the ground-truth point cloud (${\mathcal{P}}$) along with several other 2D losses on the Peeled maps, as listed in Sec. \ref{sec:loss}.}
    \label{fig:Pipeline}
    \end{figure*}

    Thus, we reformulate the solution to the monocular textured 3D body reconstruction task as predicting a set of Peeled Depth \& RGB maps. To achieve this dual-prediction task, we propose PeelGAN, a dual-task generative adversarial network that generates a set of depth and RGB maps in two different branches of the network, as shown in Figure \ref{fig:Pipeline}. These predicted peeled maps are then back-projected to 3D space to obtain a point cloud. Similar to~\cite{wang2018pixel2mesh}, we propose to include Chamfer loss over the reconstructed point cloud in the camera coordinate frame. This loss implicitly imposes a 3D body shape regularization during training. Our model is able to hallucinate plausible parts of the body that are self-occluded in the image. As compared to PIFu and MouldingNet, PeelGAN has the advantage of being computationally efficient while handling severe self-occlusions and arbitrary surface topology deformations caused by loose clothing. Our proposed representation enables an end-to-end, non-parametric and differentiable solution for textured 3D body reconstruction.
    
    We evaluate our method with prior work on public datasets such as BUFF~\cite{zhang2017detailed} and MonoPerfCap~\cite{xu2018monoperfcap}. MonoPerfCap consists of articulated skeletal motions and medium-scale non-rigid surface deformations by deforming a template mesh. Hence, loose clothing and large scale non-rigid deformations are not included. On the other hand, BUFF sequences are noisy with limited variations in shape and clothing. To compensate for the lack of realistic 3D datasets with large variations in shape and clothing, we present a challenging 3D dataset captured from our calibrated multi-Kinect setup. It consists of 8 subjects with large variations in loose clothing and shape (see Sec. \ref{sec:our_data}). We evaluate our method on all three datasets and report superior quantitative and qualitative results to other state-of-the-art methods. 
    To summarize our contributions in this paper:
    \begin{itemize}
        \item We introduce PeeledHuman - a novel shape representation of the human body encoded as a set of Peeled Depth and RGB maps, that is robust to severe self-occlusions.
        \item Our proposed representation is efficient in terms of both encoding 3D shapes as well as feed-forward time yielding superior quality of reconstructions with faster inference rates.
        \item We propose PeelGAN - a complete end-to-end pipeline to reconstruct a textured 3D human body from a single RGB image using an adversarial approach.
        \item We introduce a challenging 3D dataset consisting of multiple human action sequences with variations in shape and pose, draped in loose clothing. We intend to release this data along with our code for academic use. 
    \end{itemize}
   
\section{Related Work}
\label{sec:RW}
    Traditionally, voxel carving and triangulation methods were employed for recovering a 3D human body from calibrated multi-camera setups~\cite{dou2016fusion4d,bogo2017dynamic}. 
    Majority of existing deep learning methods to recover 3D shapes from monocular RGB images use parametric SMPL~\cite{SMPL:2015} model. HMR~\cite{hmr} proposes to regress SMPL parameters while minimizing re-projection loss. 
    Segmentation masks~\cite{varol2017learning} were used to further improve the fitting of the 3D model to the available 2D image. However, these parametric body estimation methods yield a smooth naked mesh missing out on surface geometry details. Additionally, researchers have explored to incorporate tight clothing details over the SMPL model by estimating displacements of each vertex~\cite{bhatnagar2019mgn,alldieck2019learning}. Very recently, clothing deformation is predicted as a function of garment size~\cite{tiwari2020sizer}. Authors in~\cite{Venkat_2019_ICCV_Workshops} estimate vertex displacements by regressing to SMPL vertices. These techniques fail for complex clothing topologies such as skirts and dresses.
        
    On the other hand, model-free approaches do not use any parametric model. Volumetric regression~\cite{varol18_bodynet,VenkatSS18,huang2018deep} uses a voxel grid, i.e., a binary occupancy map to recover the human body from a single RGB image. 
    Volumetric representations pose a serious computational disadvantage due to the sparsity of the voxel grid and surface quality is limited to the voxel grid resolution. 
    Deformation based approaches have been proposed over parametric models which incorporate these details to an extent. 
    The constraints from body joints, silhouettes, and per-pixel shading information are utilized in~\cite{hmd} to produce per-vertex movements away from the SMPL model. However, only the visible pixels are modeled in this approach. 
    
    To address the aforementioned issues during the reconstruction of 3D human bodies, interest has garnered around non-parametric approaches recently. 
    Deep generative models have been proposed in~\cite{natsume2019siclope} taking inspiration from the visual hull algorithm to synthesize 2D silhouettes that are back-projected from inferred 3D joints. The silhouettes are back-projected to obtain clothed models with different shape complexities.
    Implicit representations of 3D objects have been employed for deep learning-based approaches in~\cite{mescheder2019occupancy,saito2019pifu,saito2020pifuhd, Li2020portrait,bhatnagar2020combining,he2020geo,chibane2020implicit} which represent the 3D surface as the continuous decision boundary of a deep neural network classifier. PIFu has been extended to animate implicit representation in~\cite{huang2020arch}. Unsupervised estimation of implicit functions has been addressed in~\cite{liu2019learning,niemeyer2020differentiable}.
    Authors in~\cite{gabeur2019moulding} represent the human body as a mould and recover visible and hidden depth maps. Self-occlusions are not handled by these approaches as they do not impose any human body shape prior.  
    
    Similar to our peeled representation, multi-layer approaches have been used for 3D scene understanding. Layered Depth Images were proposed in~\cite{shade1998layered} for efficient rendering applications. Layer-structured 3D scene representation was proposed in~\cite{tulsiani2018layer} which performs view synthesis as a proxy task. Recently, transformer networks were proposed in~\cite{shin2019multi} to transfer features to a novel view to better recover 3D scene geometry. Nested shape layer representation was introduced in~\cite{richter2018matryoshka} to encode a 3D object efficiently.
    
\section{Proposed Method}

\subsection{Peeled Representation}
   \label{sec:problemFormulation}
    We encode a 3D human body model as a set of Peeled Depth \& RGB maps as follows. We assume the human body to be a non-convex object placed in a virtual scene. Given a virtual camera, a set of rays originating from the camera center are traced through each pixel to the 3D world. The set of first ray-intersections with the 3D body are recorded as depth map $d_1$ and RGB map $r_1$, capturing visible surface details that are nearest to the camera. 
    Subsequently, we \textit{peel} away the occlusion and extend the rays beyond the first bounce to hit the next intersecting surface. We successively record the corresponding depth and RGB values of the next layer as $d_i$ and $r_i$, respectively. We consider $4$ intersections of each ray \ie, $4$ Peeled Depth \& RGB maps to faithfully reconstruct a human body assuming this can handle self-occlusions caused by the most frequent body poses.  
   
    A point cloud can be constructed from these maps using classical camera projection methods. If the camera intrinsics, \ie, the focal length of camera $f = [f_x, f_y]$ and its center of axes $C = [C_x, C_y]$ are known, then the ray direction in the camera coordinate frame corresponding to pixel $[X,Y]$ is given as

    \begin{align}
        ray[X,Y] = \bigg(\frac{X-C_x}{f_x}, \frac{Y-C_y}{f_y}, 1 \bigg).
    \end{align}
    For a pixel $[X,Y]$ with depth $d^{XY}_{1}$ in the first depth map, its 3D location in the camera coordinate frame is given by

    \begin{align}
        \renewcommand{\arraystretch}{2}
        \begin{bmatrix}
            x\\
            y\\
            z
        \end{bmatrix}
        =
        \begin{bmatrix}
            \dfrac{X_{norm} \cdot d^{XY}_{1}}{fx}\\
            \dfrac{Y_{norm} \cdot d^{XY}_{1}}{fy}\\
            d^{XY}_{1}
        \end{bmatrix},
    \label{eq:backproject}
    \end{align}
    where $X_{norm} = X - h/2$ and $Y_{norm} = Y - w/2$. Here, we assume $[h/2, w/2]$ is the center of the image. \\
    \\
    \textbf{Problem Formulation} Given an RGB image $r_1$ of resolution $(h \times w \times 3)$ captured from an arbitrary viewpoint, our goal is to reconstruct a textured 3D body model from $n$ Peeled Depth maps ($\widehat{\mathcal{D}}$) and $n-1$ Peeled RGB maps ($\widehat{\mathcal{R}}$) where $\widehat{\mathcal{D}} = \{\hat{d}_1, \hat{d}_2, \cdots, \hat{d}_{n}\}$ and $\widehat{\mathcal{R}} = \{\hat{r}_2, \hat{r}_3, \cdots, \hat{r}_{n-1}\}$ respectively. The ground-truth maps are denoted as $\mathcal{D} = \{d_1, d_2, \cdots, d_{n}\}$ and $\mathcal{R} = \{r_1, r_2, \cdots, r_{n}\}$. A reconstructed point cloud $\widehat{\mathcal{P}}$ is obtained using Eq. \ref{eq:backproject}. The ground-truth point cloud ${\mathcal{P}}$ is used as 3D supervision in Eq. \ref{eq:chloss}. We do not generate $\hat{r}_1$ as the input image $r_1$ can be considered as the first generated RGB map. We use $n = 4$ maps in our method. Background pixels have depth value $0$ and RGB value ($255,255,255$). They do not constitute $\widehat{\mathcal{P}}$. For body poses with less than $4$ ray intersections, $d_3$ and $d_4$ are $0$ while $r_3$ and $r_4$ are equal to the background color. At test time, only pixels with predicted non-zero depth values are backprojected.

    \subsection{PeelGAN}
    \label{sec:loss}
    To generate Peeled maps from an input image, we propose  a conditional GAN, named PeelGAN, as depicted in  Figure \ref{fig:Pipeline}. PeelGAN takes a single RGB image as its input and generates Peeled Depth maps $\widehat{\mathcal{D}}$ and corresponding RGB maps $\widehat{\mathcal{R}}$ (refer to Sec. \ref{sec:problemFormulation}). 
    The input RGB image is first fed to an encoder network (similar to~\cite{isola2017image}) consisting of a few convolutional layers for recovering $128\times128\times256$ feature maps and is subsequently fed to a
    series of 18 ResNet~\cite{resnet} blocks. The network uses ReLU as its activation function. We propose to decode the Peeled Depth and RGB maps in two separate branches since they are sampled from different distributions. The network produces $3$ Peeled RGB maps and $4$ Peeled Depth maps which are then separately fed to two different discriminators, one for each RGB and Depth maps. We use PatchGAN discriminator as proposed in~\cite{isola2017image}. We denote our generator as $G$, the Peeled RGB map discriminator as $D_r$ and the Peeled depth map discriminator as $D_d$. We train our network with the following loss function:
    \begin{multline}
        \begin{aligned}
            L_{peel} = L_{gan}+ \lambda_{depth}L_{depth} + \lambda_{rgb}L_{rgb} \\
            + \lambda_{cham}L_{cham} + \lambda_{smooth}L_{smooth},
        \end{aligned}
    \label{loss}
    \end{multline}
    where $\lambda_{depth}$, $\lambda_{rgb}$, $\lambda_{cham}$, $\lambda_{smooth}$ are weights for depth loss($L_{depth}$), RGB loss($L_{rgb}$), Chamfer loss($L_{cham}$) and smoothness loss($L_{smooth}$) respectively. Each loss term is explained in detail below. \\
    \\
    \textbf{GAN Loss} ($L_{gan}$) We follow the usual GAN objective for the generated $\widehat{\mathcal{R}}$ and $\widehat{\mathcal{D}}$ maps conditioned on the input image $r_0$ as
    \begin{multline}
        \begin{aligned}
        L_{gan} = E_{r_0,\mathcal{R}}[log D_{r}(r_0, \mathcal{R})] + E_{r_0,\mathcal{D}}[log D_{d}(r_0, \mathcal{D})] \\ + E_{r_0}[log(1 - D_r(r_0, \widehat{\mathcal{R}}))] +  E_{r_0}[log(1 - D_d(r_0,\widehat{\mathcal{D}}))].
        \end{aligned}
    \label{eq:state-space&obs-equ}
    \end{multline}
    \\
    \textbf{Depth Loss} ($L_{depth}$)
    We minimize the masked L1 loss over ground-truth and generated peeled depth maps. $\gamma$ is used as a weighting factor to encourage prediction of self-occluded parts appearing in $d_3$ and $d_4$ as 
    \begin{equation}
        L_{{depth}} = \sum_{i=1}^{4} \Big\lVert{m_i \cdot (d_i - \hat{d}_{i})}\Big\Vert_1,
    \end{equation}
    where $m_i=\gamma$ ($>$1) for occluded pixels and $m_i = 1$ otherwise. \\
    \\
    \textbf{RGB Loss} ($L_{rgb}$) The generator minimizes L1 loss between the ground-truth and generated peeled RGB maps as \\
    \begin{align}
        L_{rgb} = \sum_{i=2}^{4} \Big\lVert{(r_i - \hat r_{i})}\Big\Vert_1.
    \end{align}
    \\
    \textbf{Chamfer Loss} ($L_{cham}$) To enable the network to capture the underlying 3D structure of the generated depth maps, we minimize Chamfer distance between the reconstructed point cloud ($\widehat{\mathcal{P}}$) and the ground-truth point cloud ($\mathcal{P}$),
    \begin{equation}
    \begin{aligned}
        L_{cham}(\widehat{\mathcal{P}}, \mathcal{P}) &= \sum_{\Vec{p_i} \in \widehat{\mathcal{P}}} \min_{\vec{q_j} \in \mathcal{P}} \lVert{\Vec{p_i} - \Vec{q_j}} \Vert^2_2 + \sum_{\Vec{q_j} \in \mathcal{P}} \min_{\Vec{p_i} \in \widehat{\mathcal{P}}} \lVert{\Vec{q_j} - \Vec{p_i}} \Vert^2_2.
    \end{aligned}    
    \label{eq:chloss}
    \end{equation} 
    Chamfer loss induces 3D supervision by fusing multiple independent 2.5D generated peel depth maps. Refer Sec. \ref{ablation} for evaluation of Chamfer loss. \\
    \\
    \textbf{Smoothness Loss} ($L_{smooth}$) There is an additional need to enforce smoothness in depth variations over the surface (except for the boundary regions). Thus, motivated by~\cite{tan2020self}, we enforce the first derivative of generated Peeled Depth maps to be close to that of the ground-truth Peeled Depth maps as
    \begin{equation}
        \begin{aligned}
            L_{{smooth}} = \sum_{i=1}^{4} \Big\lVert{\bigtriangledown d_i - \bigtriangledown \hat{d}_{i}}\Big\Vert_1 
        \end{aligned}
    \label{eq:smoothloss}
    \end{equation}
\section{Experiments}
\label{sec:Exp}

\subsection{Datasets and Preprocessing}
\label{sec:our_data}
We perform qualitative and quantitative evaluation on three datasets, namely (i) BUFF~\cite{zhang2017detailed} (ii) MonoPerfCap~\cite{xu2018monoperfcap} (iii) Our new dataset. We scale each 3D body model to a unit-box and compute $4$ Peeled Depth and RGB maps from $4$ different camera angles each: $0^{\circ}$ (canonical view), $45^{\circ}$, $60^{\circ}$, $90^{\circ}$. \\
\\
\textbf{BUFF Dataset} consists of $5$ subjects with tight and loose clothing performing complex motions. The dataset consists of 11,054 3D human body models in total. We use this completely for testing our method. \\
\\
\textbf{MonoPerfCap Dataset} consists of 13 daily human motion sequences in tight and loose clothing styles. It has approximately 40,000 3D human body models with subjects in indoor and outdoor settings. We use two sequences for inference and six sequences for training. One sequence is divided equally between training and inference. \\
\\
\textbf{Our Data} 
We introduce a 3D dataset consisting of 2,000 human body models from 8 human action sequences including marching and swinging limbs using a calibrated setup of $4$ Kinect sensors. 
The RGBD data is back-projected to obtain a point cloud and post-processed using Poisson surface reconstruction to obtain the corresponding meshes. As our data is independently reconstructed in each frame without any template constraint, we were able to capture realistic large scale deformations. The dataset contains significant variations in shape and clothing consisting of both loose and tight clothing\footnote{\textcolor{magenta}{cvit.iiit.ac.in/research/projects/cvit-projects/3dcomputervision}}. We use six sequences for training and two sequences for inference. The dataset will be released for academic purposes to spur further research in this field.

\subsection{Training Protocol}
We implement our proposed pipeline in PyTorch using 4 Nvidia GTX 1080 Ti GPUs with 11GB RAM trained for 45 epochs. A batch size of $12$ is used for $512 \times 512$ images. Ground-truth Peeled maps are captured using trimesh \footnote{\textcolor{magenta}{trimsh.org}}. We use the Adam optimizer with a learning rate of $1.5e$-$4$ and $\gamma$, $\lambda_{dep}$, $\lambda_{cham}$, $\lambda_{rgb}$ and $\lambda_{smooth}$ as 10, 100, 500, 500, 500, respectively. One sequence from the MonoPerfCap dataset was used as validation set for grid search over all hyperparameters. The final predicted point cloud contains $30000$ 3D body surface points on average. 


\begin{figure*}[t!]
\begin{center}
    \includegraphics[width=0.8\linewidth]{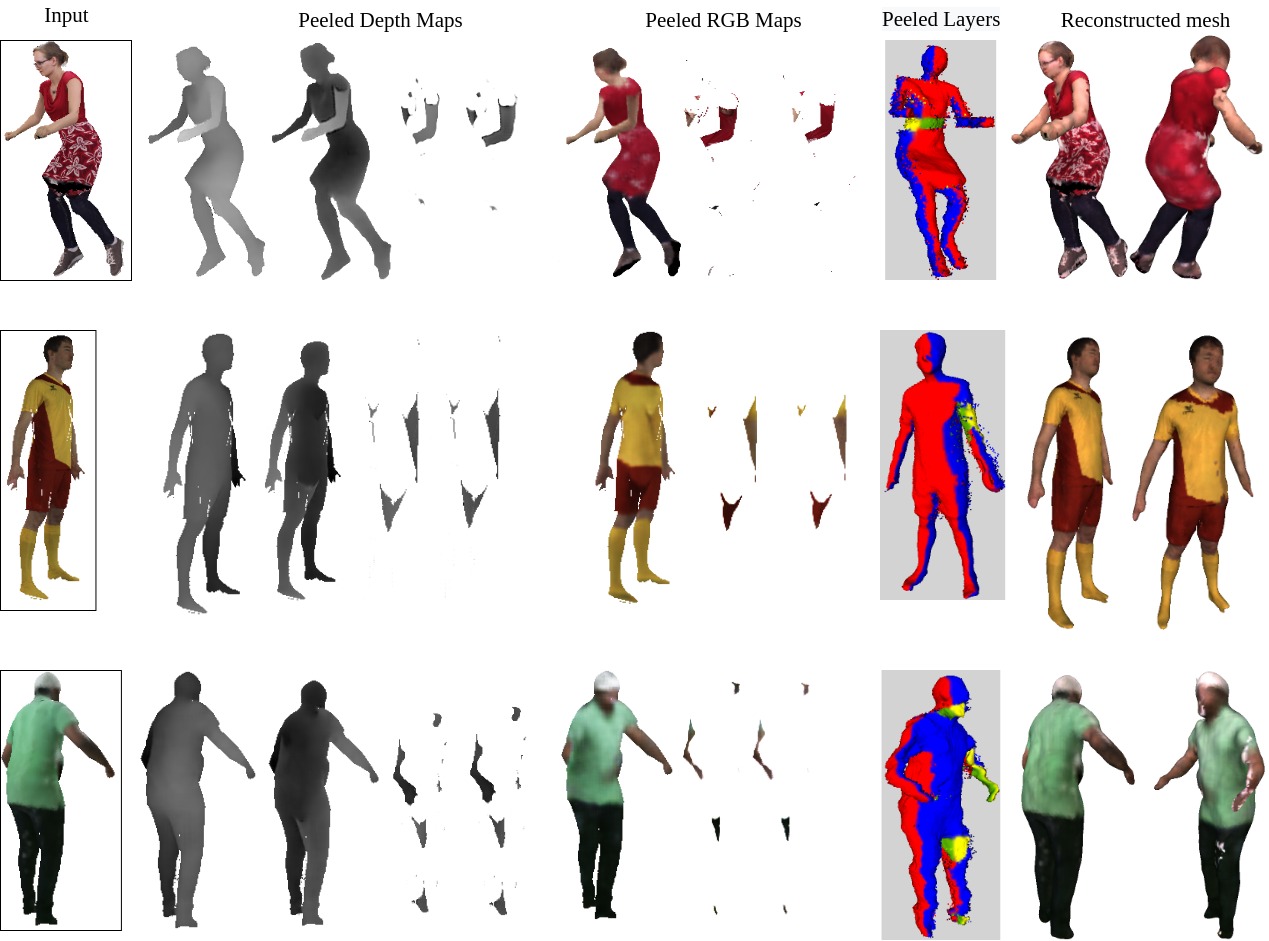}
\end{center}
\caption{\textbf{Qualitative results} on MonoPerfCap (Top row), BUFF (Middle row), and Our Dataset (Bottom row). For each subject, we show (from left to right) input image, $4$ Peeled Depth and RGB maps, backprojected Peeled layers (colored according to their depth order : red, blue, green, and yellow respectively), reconstructed textured mesh. Please refer to the supplementary material for an extended set of results.}
\label{fig:illusresults}
\end{figure*}

\begin{figure*}
\begin{center}
    \includegraphics[width=0.8\linewidth]{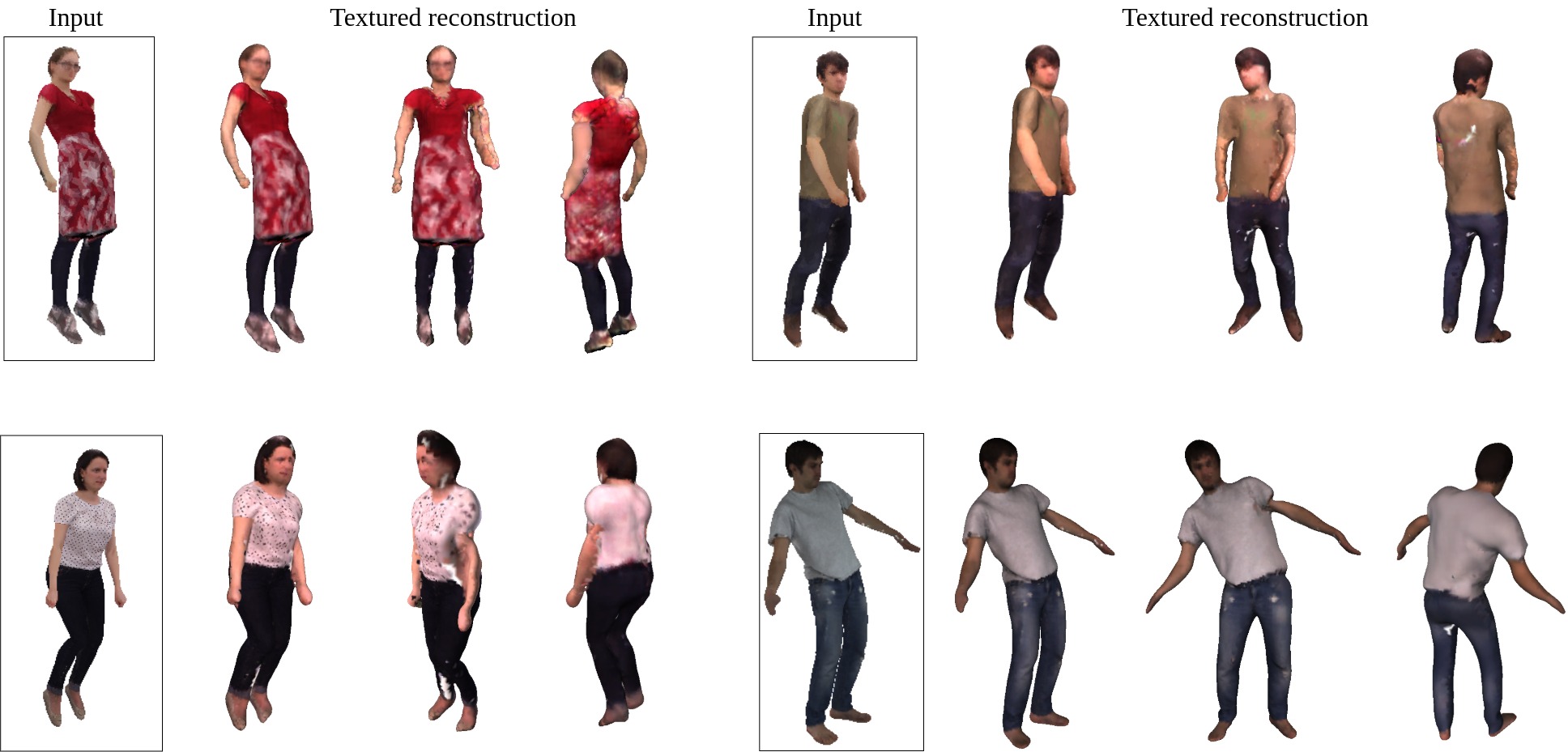}
\end{center}
\caption{\textbf{Qualitative textured reconstruction results} on MonoPerfCap and BUFF datasets. For each subject, we show the input image and multiple views of the reconstructed mesh (after performing Poisson surface reconstruction on the reconstructed point cloud). Our proposed PeeledHuman representation efficiently reconstructs the occluded parts of the body from a single view.}
\label{fig:multiviewresults}
\end{figure*}

\begin{figure*}[t!]
  \includegraphics[width=\linewidth]{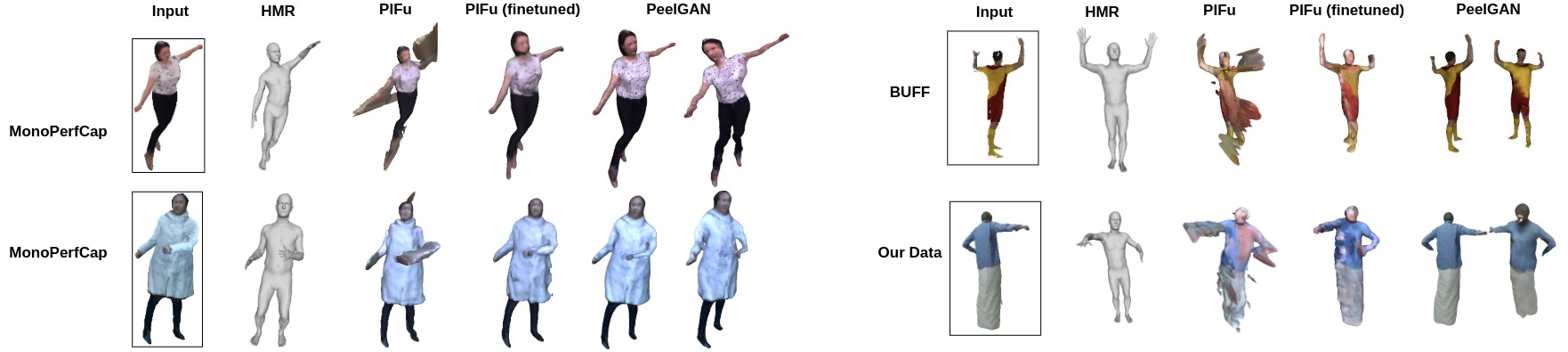}
  \caption{\textbf{Qualitative comparison of HMR and PIFu with PeelGAN} for MonoPerfCap, BUFF, and Our Dataset. Our method can reconstruct plausible shapes efficiently even under severe self-occlusions.}
  \label{fig:Comparison}
\end{figure*}

\subsection{Qualitative Results}
We demonstrate single-view/monocular reconstruction results on all $3$ datasets in Figure \ref{fig:illusresults} and Figure \ref{fig:multiviewresults}. Our method can accurately recover the 3D human shape from previously unseen views. Due to the nature of our encoding, our method can recover the self-occluded body parts reasonably well for severely occluded views.

\begin{figure}
\centering
  \begin{subfigure}{\linewidth}
    \centering
    \includegraphics[width=\linewidth]{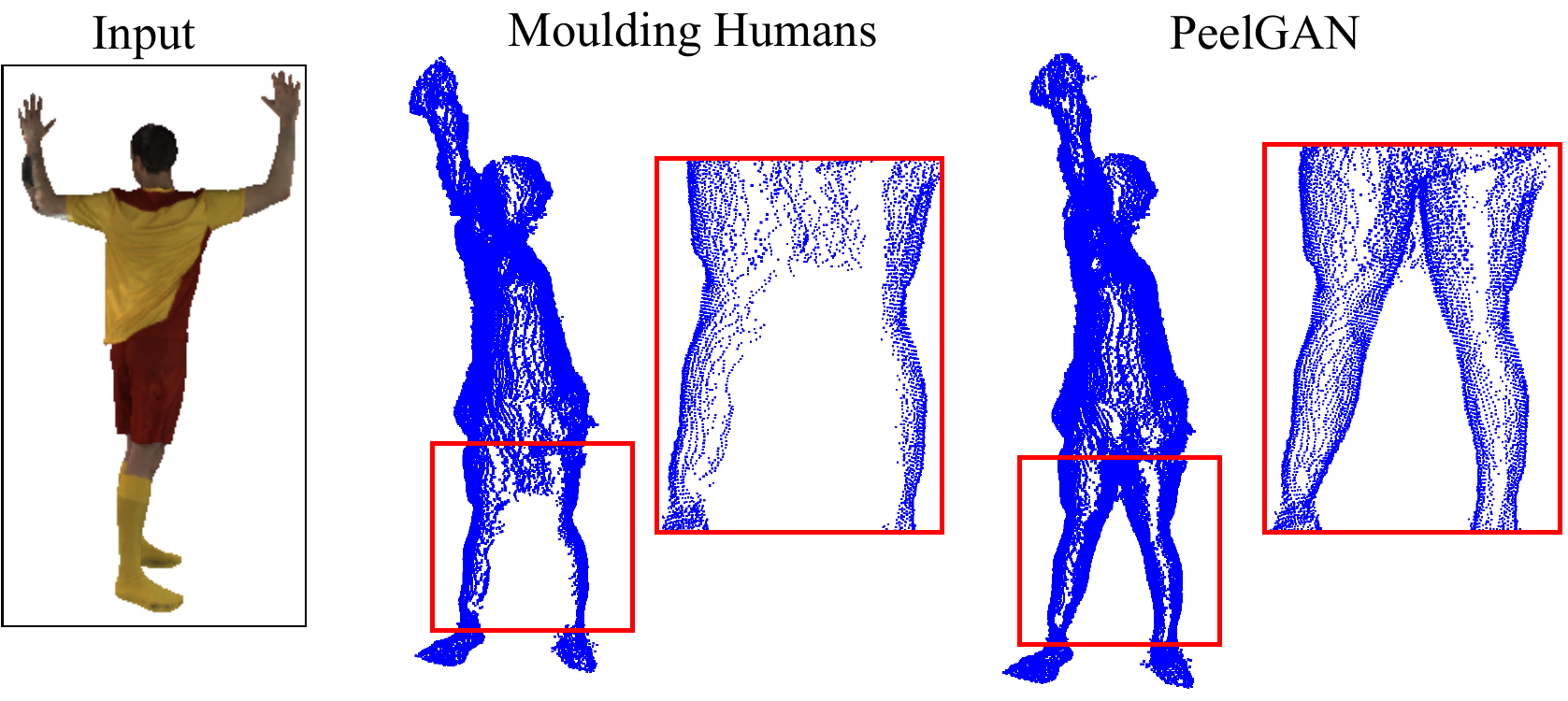}
    \caption{}
    \label{fig:moulding}
  \end{subfigure}
  \begin{subfigure}{\linewidth}
    \centering
    \includegraphics[width=0.98\linewidth]{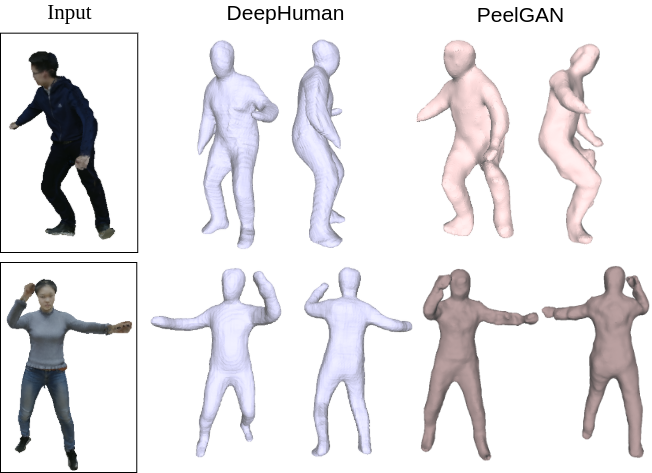}
    \caption{}
    \label{fig:thumans}
  \end{subfigure}
 \caption{Qualitative comparison with (a) Moulding Humans~\cite{gabeur2019moulding} (trained on MonoPerfCap and our dataset) (b) DeepHuman~\cite{zheng2019deephuman} (trained on THUman dataset). Both methods fail to recover the shape and surface texture accurately.}
\label{fig:mouldanddeephuman}
\end{figure}

\subsection{Comparison with Prior Work}
We perform qualitative comparison of our proposed representation with other commonly used representations for single-view 3D human reconstruction. In particular, we compare our method with parametric body model regression (meshes) and implicit function learning methods in Figure \ref{fig:Comparison} as well as, with voxel regression and point cloud regression method in Figure \ref{fig:mouldanddeephuman}. We retrain PIFu~\cite{saito2019pifu} using MonoPerfCap and our dataset. We also evaluate PIFu after finetuning the model provided by authors with MonoPerfCap and our dataset. We compare with HMR~\cite{hmr} as a parametric model regression (mesh-based) method. To compare against MouldingNet~\cite{gabeur2019moulding} in Figure \ref{fig:moulding}, we train PeelGAN with two depth maps and our own specifications as neither code nor data was made public by the authors. For voxel-based method, we train PeelGAN model and DeepHuman~\cite{zheng2019deephuman} (predicts only textureless models) using the released THUman dataset~\cite{zheng2019deephuman} shown in Figure \ref{fig:thumans}. 
\begin{table}[tbh!]
\centering
\begin{tabular}{lcc}
\toprule
Method & Chamfer Distance $\downarrow$ & Image Resolution\\
\midrule
    BodyNet~\cite{varol18_bodynet} & 4.52 & 256\\
    SiCloPe~\cite{natsume2019siclope} & 4.02 & 256\\
    VRN~\cite{jackson20183d} & 2.48 & 256\\
    PIFu~\cite{saito2019pifu} & 1.14 & 512\\
    \hline
    Ours & 1.283 & 256\\
    \textbf{Ours} & \textbf{0.9254}&\textbf{512}\\
\bottomrule
\end{tabular}
\caption{Quantitative comparison with other methods. Our method achieves the lowest Chamfer score for single-view reconstruction, indicating the robustness of our representation.}
\label{quant}
\end{table}%

As demonstrated in Figure \ref{fig:Comparison}, our proposed method consistently recovers the underlying shape and texture. When trained from scratch, PIFu fails to recover shape but finetuning the pre-trained model (trained on commercial high-resolution meshes) results in lesser artifacts. This emphasizes the necessity of high-resolution data to train implicit function approaches. Moreover, PIFu is not end-to-end trainable since it requires to train shape and color components separately. HMR produces a smooth naked body mesh missing surface texture details.
MouldingNet fails to recover body shape when there is significant self-occlusion in the input image, as seen in Figure \ref{fig:moulding}. Our method recovers plausible human shapes even when it is challenging to distinguish body parts from a single-view as shown in Figure \ref{fig:thumans} (here hand is indistinguishable from torso due to textureless dark-shaded clothing). 

Quantitative evaluation of our method using Chamfer distance against PIFu, BodyNet~\cite{varol18_bodynet}, SiCloPe~\cite{natsume2019siclope} and VRN~\cite{jackson20183d} is shown in Table \ref{quant}. Here we report results on both $512$ resolution and $256$ resolution inputs to have a fair comparison with other methods. We can conclude that our method achieves significantly lower Chamfer distance values as compare to other existing methods. 

\subsection{Discussion}
\subsubsection{Ablation Study}\label{ablation}
\begin{figure}
\centering
  \begin{subfigure}{\linewidth}
    \centering
    \includegraphics[width=\linewidth]{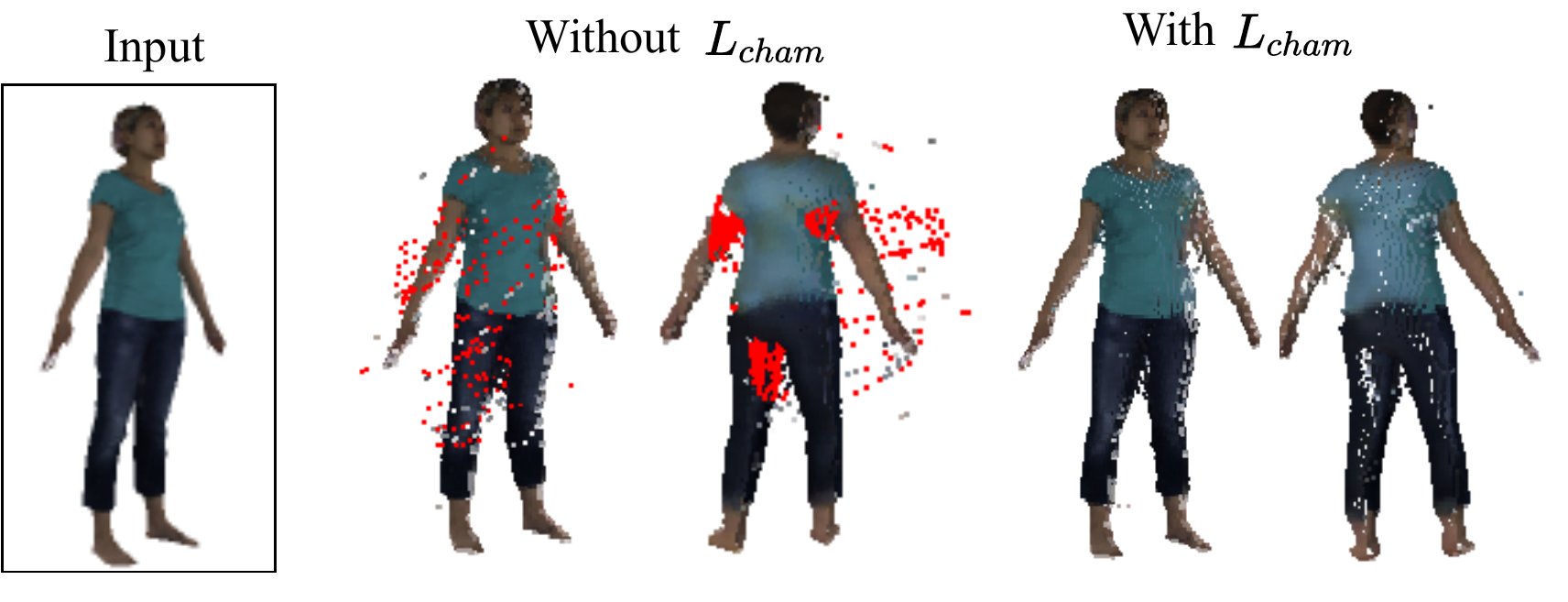}
    \caption{}
    \label{fig:chamfer}
  \end{subfigure}
  \begin{subfigure}{\linewidth}
    \centering
    \includegraphics[width=\linewidth]{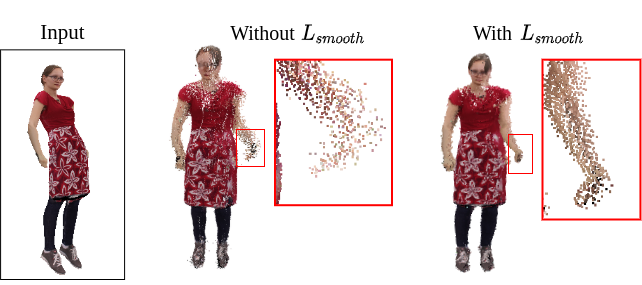}
    \caption{}
    \label{fig:smoothness}
  \end{subfigure}
 \caption{(a) Reconstruction without and with Chamfer loss. Red points indicate both noise and occluded regions that were not predicted by the network. (b) Training with smoothness loss improves the quality of Peeled Depth maps.}
\label{fig:smoothness_chamfer}
\end{figure}

 We perform a few ablative studies to demonstrate the effect of Chamfer and smoothness losses on the reconstruction quality of our method. Firstly, we train our network without Chamfer loss. The network is not able to hallucinate the presence of occluded parts in the $3^{rd}$ and $4^{th}$ depth maps and are hence missing in Figure \ref{fig:chamfer}. We also observe that absence of Chamfer loss produces significant noise in reconstructions (red points). This can be attributed to independent predictions of individual depth maps using L1 loss. 
 \newline
 We also study the effect of smoothness loss (Eq. \ref{eq:smoothloss}). This helps the network to produce smoother depth values in layers as shown in Figure \ref{fig:smoothness}. Thus, Chamfer loss forces the network to predict plausible shapes, that are often noisy, for the occluded parts. Smoothness loss helps the network to smooth out these noisy depth predictions. 
 
\begin{figure}
    \centering
    \includegraphics[width=\linewidth]{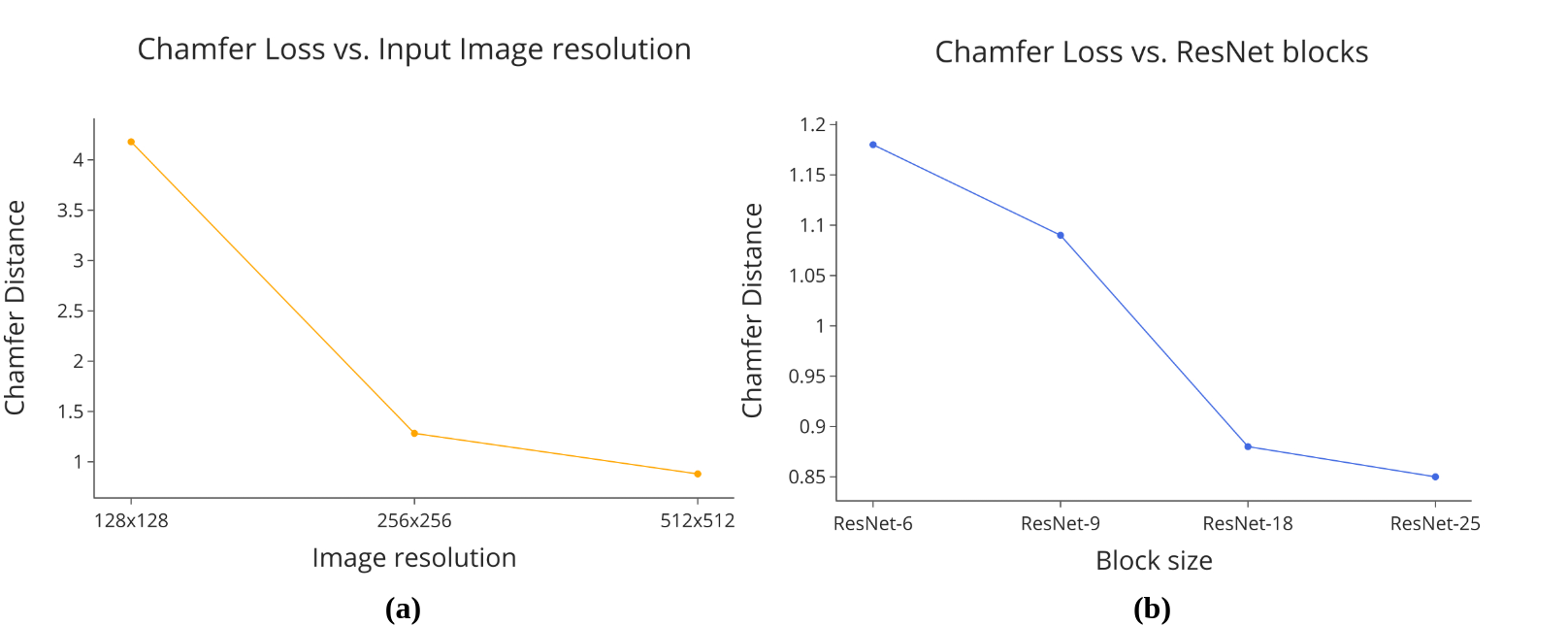}
    \caption{(a) Chamfer loss vs. Input image resolution (b) Chamfer loss vs. ResNet blocks}
\label{fig:Plots}
\end{figure}

\subsubsection{In-the-wild images}
\begin{figure}
    \centering
    \includegraphics[width=\linewidth]{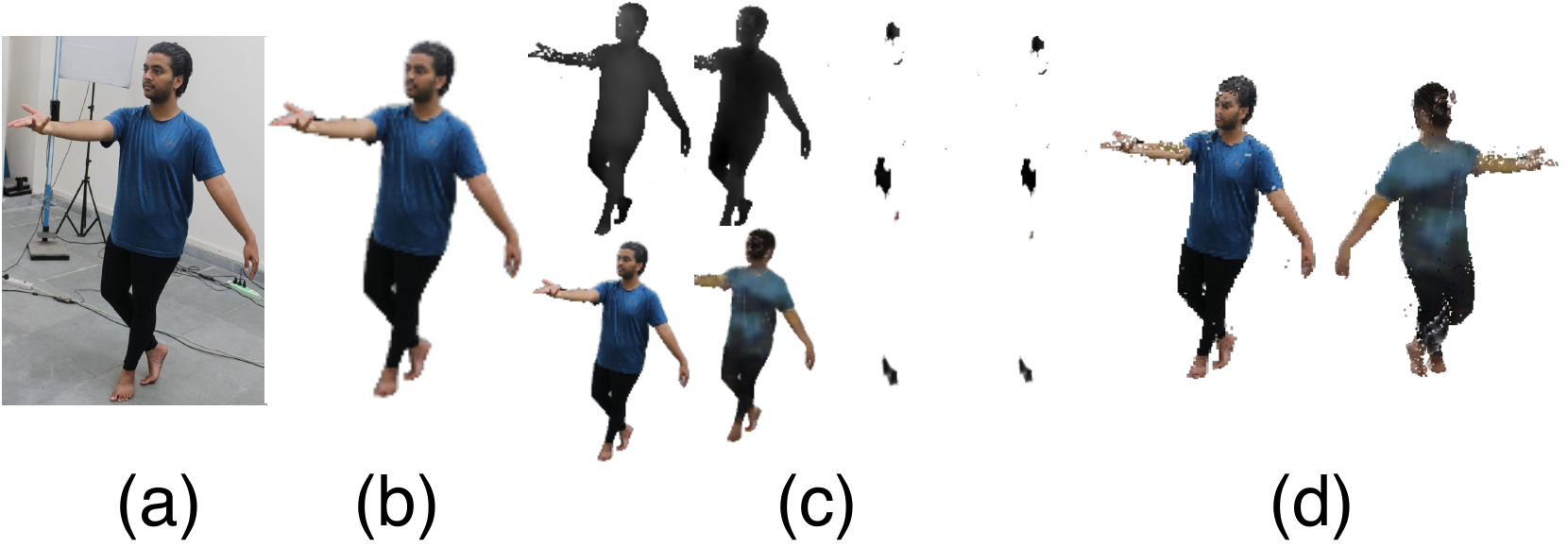}
    \caption{Performance of our method on in-the-wild images.}
    \label{fig:Supreeth&nonoise}
  \end{figure}

 We also showcase results in Figure \ref{fig:Supreeth&nonoise} on an in-the-wild image not present in any dataset. We segment the input image using~\cite{gong2019graphonomy} before feeding it to our model. The predicted Peeled Depth and RGB maps are visualized in (c) and final textured reconstruction in (d). This shows that our method can handle wide varieties in shape, pose, and texture.

\subsubsection{Effect of Input Resolution and ResNet blocks}
    We demonstrate the effect of ResNet blocks and input image resolutions on the performance of PeelGAN in Figure \ref{fig:Plots}. As we can observe, Chamfer loss decreases with an increase in input image resolution. A similar trend is observed for increasing the number of ResNet blocks. Since the improvement in Chamfer loss from ResNet-18 to ResNet-25 is not significant, we stick to using ResNet-18 for our experiments. 
\section{Conclusion}
    We present a novel representation to reconstruct a textured human model from a single RGB image using Peeled Depth and RGB maps. Such an encoding is robust to severe self-occlusions while being accurate and efficient at learning \& inference time.
    Our peeled representation miss to capture few surface triangles that are tangential to the viewpoint of the input image. However, this limitation can be addressed with minimal post-processing when constructing meshes from the corresponding predicted point clouds. 

{\small
\bibliographystyle{ieee}
\bibliography{egpaper_final}
}
\end{document}